\documentclass[sigconf]{acmart}
\usepackage{graphicx} 
\usepackage{multicol}
\usepackage{amsmath}
\usepackage{enumitem}
\usepackage{float}
\usepackage{bm}
\usepackage{url}
\usepackage{siunitx}

\AtBeginDocument{%
  \providecommand\BibTeX{{%
    \normalfont B\kern-0.5em{\scshape i\kern-0.25em b}\kern-0.8em\TeX}}}

\acmConference[2024]{}{}{}

\begin{document}
\title{FastLogAD: Log Anomaly Detection with Mask-Guided Pseudo Anomaly Generation and Discrimination}

\author{Yifei Lin}
\email{ylin20@ualberta.ca}
\affiliation{%
  \institution{University of Alberta}
  \city{Edmonton}
  \state{Alberta}
  \country{Canada}
}

\author{Hanqiu Deng}
\email{hanqiu1@ualberta.ca}
\affiliation{%
  \institution{University of Alberta}
  \city{Edmonton}
  \state{Alberta}
  \country{Canada}
}

\author{Xingyu Li}
\email{xingyu@ualberta.ca}
\affiliation{%
  \institution{University of Alberta}
  \city{Edmonton}
  \state{Alberta}
  \country{Canada}
}


\begin{abstract}
Nowadays large computers extensively output logs to record the runtime status and it has become crucial to identify any suspicious or malicious activities from the information provided by the real-time logs. Thus, fast log anomaly detection is a necessary task to be implemented for automating the infeasible manual detection. Most of the existing unsupervised methods are trained only on normal log data, but they usually require either additional abnormal data for hyperparameter selection or auxiliary datasets for discriminative model optimization. In this paper, aiming for a highly effective discriminative model that enables rapid anomaly detection, we propose FastLogAD, a generator-discriminator framework trained to exhibit the capability of generating pseudo-abnormal logs through the Mask-Guided Anomaly Generation (MGAG) model and efficiently identifying the anomalous logs via the Discriminative Abnormality Separation (DAS) model. Particularly, pseudo-abnormal logs are generated by replacing randomly masked tokens in a normal sequence with unlikely candidates. During the discriminative stage, FastLogAD learns a distinct separation between normal and pseudo-abnormal samples based on their embedding norms, allowing the selection of a threshold without exposure to any test data and achieving competitive performance. Extensive experiments on several common benchmarks show that our proposed FastLogAD outperforms existing anomaly detection approaches. Furthermore, compared to previous methods, FastLogAD achieves at least x10 speed increase in anomaly detection over prior work. Our implementation is available at \url{https://github.com/YifeiLin0226/FastLogAD}.
\end{abstract}





\maketitle

\section{Introduction}
\begin{figure}[ht]
    \centering
    \includegraphics[scale = 0.55]{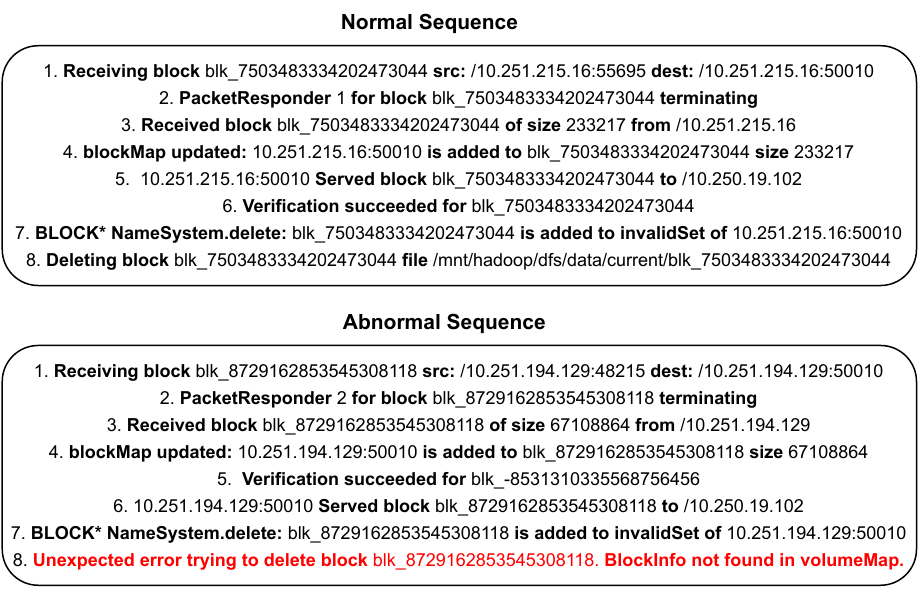}
    \caption{Examples of normal and abnormal log sequences.}
    \label{fig:sample-log}
\end{figure}

Anomaly detection is a fundamental task in machine learning, focusing on the identification of deviations from expected patterns within a dataset, especially when confronted with limited or no prior instances of anomalies. In the context of large-scale computer systems, logs record details of system operations by capturing real-time interactions between data, users and files. While the majority of log entries document routine system activities, certain log sequences exhibit deviations from the norm, signifying abnormal operations that have the potential to result in performance degradation or security threats. To illustrate this concept, Figure \ref{fig:sample-log} presents a fragment comprising both normal and abnormal log sequences. In the first seven lines, these sequences adhere to identical log templates, representing typical and expected system behavior. Besides, a conspicuous deviation manifests in the last line of the abnormal sequence, indicating a potential error. In practice, real-time log sequences are more extensive and intricate than this simplified example, rendering manual anomaly detection an challenging endeavor due to the sheer volume and complexity of data involved.

It should be noted that in the realm of system maintenance and security, the speed of log anomaly detection plays a pivotal role. As systems grow increasingly complex and voluminous, the ability to swiftly process and analyze logs at high throughput becomes critical. Fast detection not only ensures timely identification of potential issues, thereby reducing downtime and mitigating risks, but also enhances the overall resilience of systems against emerging threats. In this light, optimizing for speed in log anomaly detection is not just a technical necessity; it is a strategic imperative that underpins the security and reliability of modern computing environments. Thus beyond log anomaly detection accuracy, this paper aims to address the crucial need for speed in log anomaly detection and focuses on investigating innovative approaches that significantly improve detection throughput without compromising accuracy.

Among various learning-based solutions proposed to tackle the task of log anomaly detection, most studies primarily focus on the unsupervised scenario where the training data consists only of normal instances and can be categorized into those based on discriminative tasks \cite{zhang2019robust,nedelkoski2020self,yang2021semi,wittkopp2021a2log,xie2022loggd} and log language generative tasks \cite{du2017deeplog,meng2019loganomaly,guo2021logbert,wan2021glad,jia2021logflash,qi2023loggpt}. The former directly optimizes a binary classifier for log anomaly detection and its fast inference speed is a significant advantage. However, with only task-specific normal data available for model training, previous methods often used extra data from other sources to act as abnormal logs for discriminator training \cite{nedelkoski2020self,yang2021semi,wittkopp2021a2log}. However, these extra data may not be representative of the target domain, leading to degraded performance in model deployment. The latter based on log language generative models does not rely on additional data and achieve anomaly detection by modeling the sequential pattern of normal logs in model training. Such solutions usually first train a generative model to predict the next or masked entries in normal log sequences. Then, anomalies are detected by examining if the target log entry is in the top-K list predicted by the generative model. Due to the gap in the objective between the training and testing phases under the paradigm of log language generative modeling \cite{qi2023loggpt}, extensive abnormal data are usually required for the optimal hyperparameter selection in downstreaming anomaly detection strategies. In addition, these generative models, while powerful, tend to be complex and computationally intensive due to their inherent regression property, leading to slower inference speeds.

In response to these challenges, we propose FastLogAD for fast log anomaly detection. Similar to the architecture of the ELECTRA \cite{clark2020electra}, our FastLogAD model consists of a Mask-Guided Anomaly Generator (MGAG) and log abnormality discriminative network. The former essentially is a transformer-based log language generative model, aiming to learn the sequential patterns in normal logs. Then two novel sampling strategies, namely random replacing and masked token replacing, are introduced and the generative model is used in a innovated manner to generate pseudo anomaly logs deviating from the normal log sequential patterns. 
Subsequently, we use a discriminator to discriminate between normal sequences and pseudo anomalous sequences, termed Discriminative Abnormality Separation (DAS). By experiencing the auxiliary pseudo-anomalies, our model can separate anomalies by enforcing unbounded output norms for abnormal data and keeping normal data's output norms close to the origin. This large discrepancy enables setting the threshold to the extreme quantile of output norms of normal instances, even with no exposure to anomalous instances. Additionally, the output norm corresponding to global semantic $[$CLS$]$ token for the entire sequence pattern ensures the norm is calculated for comprehensive anomaly detection. 
At the inference stage, we only use FastLogAD's discriminator to real-time detect anomaly logs, which is verified to be efficient. To the best of our knowledge, this study constitutes the first attempt to develop a one-class discriminative model for log anomaly detection that relies solely on normal logs sourced directly from the target domain.
Our experiments on HDFS \cite{xu2009detecting}, BGL \cite{oliner2007supercomputers}, and Thunderbird \cite{oliner2007supercomputers} datasets show that FastLogAD not only outperforms existing methods with the highest F1-scores, but also achieves at least x10 speed increase in anomaly detection over prior work.
The contribution of this study can be summarized in the following points:
\begin{itemize}[parsep=0pt,topsep=0pt]
    \item We propose FastLogAD, a novel approach for unsupervised log anomaly detection characterized by Mask-Guided Anomaly Generation for generating anomalies and by Discriminative Abnormality Separation for discriminating anomalies.
    \item We introduce and analyze two strategies for synthesizing pseudo-anomalies: Random Generation and Masked Language Modeling generation.
    \item We propose Replaced Token Detection and Hyperspherical Separation Training to train an anomaly discriminator and a compact hyperspherical boundary of normal features.
    \item We evaluate the proposed FastLogAD on three real anomaly detection datasets. The experiments show that FastLogAD outperform other methods and achieve real-time log anomaly detection.
\end{itemize}

\section{Related Work}

\subsection{Log Data Preprocessing}
\begin{figure*}[htbp]
    \centering
    \includegraphics[scale = 0.38]{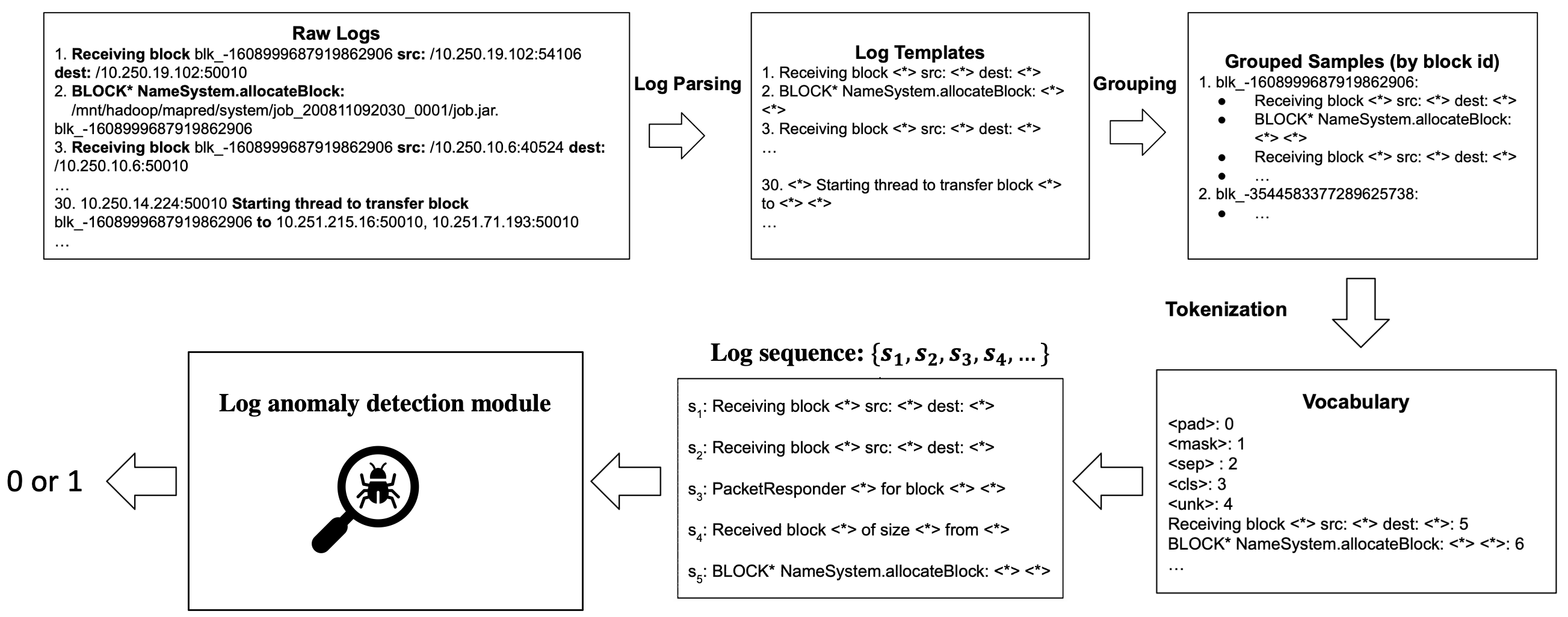}
    \caption{Illustration of the complete pipeline of the proposed log anomaly detection solution. Taking logs from the Hadoop Distributed File System (HDFS) dataset \cite{xu2009detecting} as examples, log templates are extracted through log parsing, followed by the grouping into sequences based on the identifier block ids. 
    A vocabulary is created to map the log events and special tokens (e.g., [cls], [mask]) to their corresponding unique indices during model training. Then the normal log training data is passed to the log anomaly detection module for model optimization. During inference, the vocabulary is static and used to construct query log sequences after log parsing and grouping. A yes/no answer is provided by the log anomaly detection model. The specific architecture of our log anomaly detection module is presented in Fig. 3 and Fig. 4.}
    \label{fig:processing}
\end{figure*}

Logs are often generated in unstructured or unstructured formats. Transforming raw log data into an organized and structured format is crucial for effective log analysis. Therefore, although it falls outside the primary focus of this paper, we provide a concise overview of data preprocessing such as log parsing, grouping, and tokenization, which are foundational for subsequent anomaly detection in log data.
\textbf{Log Parsing} is the process of transforming each log message into its associated string template. Logs contain both constant and variable sections, and those with matching templates are deemed to represent the same type of log event. This step is crucial in log anomaly detection as it enables the comparison of log event sequences by their order. Following log parsing, \textbf{Grouping} is performed, where parsed log events are organized into groups within a window to highlight their time-based relationships. Various grouping strategies, including fixed, sliding, and session window approaches, have been developed. For the purpose of performing quantitative analysis using deep learning techniques, \textbf{Tokenization} is applied to the parsed logs. This involves decomposing structured log events into smaller segments or tokens, with each token being assigned a unique numerical identifier. In addition to features extracted from log tokens, some approaches also incorporate extra quantitative features like count vectors \cite{meng2019loganomaly},
semantic vectors \cite{nedelkoski2020self, le2021log, li2023glad, lee2023lanobert}, and trace information \cite{zhang2022deeptralog}. 
Alternatives, there are approaches choosing not to rely on parsing due to parsing errors or Out Of Vocabulary (OOV) issues \cite{le2021log, lee2023lanobert}. They use raw log messages directly for grouping and tokenizing with, for example, WordPiece \cite{wu2016google}. While this retains the raw message and preserves semantic information, it also retains redundant semantic information, and a vocabulary of excessive size can pose challenges in training a model effectively. Thus, in this study, we follow the convention and preprocess raw logs with parsing, grouping and tokenization, which is illustrated in Fig.\ref{fig:processing}.

\subsection{Log Anomaly Detection Models}
Early studies usually utilize statistical models, for example, PCA\cite{xu2009detecting}, one-class SVM \cite{wang2004anomaly}, and Log Clustering \cite{lin2016log}, for log anomaly detection. However, they are weak in capturing log sequential patterns. So this section mainly focuses on those methods based on deep learning for sequential pattern analysis, highlighting how our study addresses the gap in the existing literature. 

\noindent\textbf{- Methods Based on Log Language Generative Models.}
In this paradigm, a deep generative model is leveraged to learn the sequential patterns within normal execution logs and identify anomalies when deviations from the established patterns occur. DeepLog \cite{du2017deeplog} stands as the pioneer, with a Long-Short-Term-Memory (LSTM) \cite{graves2012long} neural network as the backbone, and targets a model capable of predicting future tokens. Following that, LogAnomaly \cite{meng2019loganomaly} further incorporates semantic and quantitative information from log sequences. To tackle the potential long dependency issues in logs, LogBERT \cite{guo2021logbert} adopts Bidirectional Encoder Representations from Transformers (BERT) \cite{devlin2018bert} as the backbone and optimizes the model with two self-supervised tasks: masked log token prediction and volume of hypersphere minimization. Alternatively, GlAD-PAW \cite{wan2021glad} constructs a graph representation for normal log sequences and exploits the exceptional ability of Graph Neural Networks (GNNs) to capture relational dependencies among log keys for top-k next-event prediction. Recently, LogGPT \cite{qi2023loggpt} is proposed to predict the next log token. Incorporating reinforcement learning for prompt generation, it relies on ChatGPT's language understanding capabilities to analyze log messages. Despite the various designs in generative backbones and learning targets, during inference, these methods usually define a normal sequence based on the predicted output falling within a specified range of possible k logs. Nonetheless, the limitation of these approaches is prominent: The value of hyperparameter k cannot be easily selected unless abnormal data is provided. Therefore, the reported performance in literature is usually achieved by selecting the best k corresponding to the test data, which is unrealistic in a unsupervised setting where abnormal data should not be accessible before model deployment.


\noindent\textbf{- Methods Based on Discriminative Models:} LogRobust \cite{zhang2019robust} and GNN-based LogGD \cite{xie2022loggd} propose to train a binary classifier directly. To encourage compact normal feature clusters in the latent space, hyper-spherical learning-based loss originating from deep SVDD \cite{ruff2018deep} often replaces the standard cross-entropy loss in model optimization. However, due to potential unknown anomalies, the generalizability of these supervised methods on new logs is in question. In the unsupervised scenario where only normal logs are accessible, solutions including PLELog \cite{yang2021semi}, Logsy \cite{nedelkoski2020self}, A2log \cite{wittkopp2021a2log} take extra normal log data from other sources as pseudo anomalies for discriminative model optimization. However, these auxiliary data may not be representative of the target domain, leading to degraded performance in model deployment. 

In this study, aiming for a highly effective discriminative model that enables rapid anomaly detection, we employ a log language generative model to capture normal sequential patterns in logs. We then apply masked-guided sampling techniques to this model to generate synthetic abnormal logs that diverge from these learned patterns, which aids in optimizing the discriminative model. As the discriminator alone is utilized during inference, our proposed solution facilitates swift log analysis with high throughput. To our knowledge, this study represents the pioneering attempt to develop a one-class discriminative model for log anomaly detection that relies solely on normal logs sourced directly from the target domain. 

\section{Methodology}
Before elaborating the proposed log anomaly detection model, we first describe our data preparation for structured log analysis. Problem formulation and system overview are also presented.

\textbf{Log preprocessing.} In this study, we adhere to the standard practice of preparing raw logs through parsing, grouping, and tokenization, as illustrated in Fig.\ref{fig:processing}. Specifically, we employ the Drain parser \cite{he2017drain} , which uses a fixed-depth parse tree to assess the similarity between each log message and existing log events, thereby determining whether to categorize the message under an existing event or to formulate a new template. Subsequently, based on the presence of block identifiers in the logs from the target domain, we choose between session or sliding window strategies for log grouping. A vocabulary is then developed from the training log data, and all parsed logs are tokenized using this established vocabulary.

\textbf{Problem formulation.} Let $\bm{s} = \{s_1, s_2, \dots, s_d\}$ be a sequence of d log events parsed from raw log messages, following the processing steps as shown in Fig. Fig.\ref{fig:processing}. Our goal is to train a model with training data $\mathcal{D}_{train} = \{(\bm{s}_1, y_1 = 0), (\bm{s}_2, y_2 = 0), \dots, (\bm{s}_N, y_N = 0)\}$, each $y_i= 0$ represents training set consisting of only normal sequences under the unsupervised anomaly detection setting. In the subsequent detection stage, the model is able to distinguish between normal and abnormal sequences in the test set $\mathcal{D}_{test} = \{(\bm{s}_{N +1}, y_{N +1}), \dots, (\bm{s}_{N + M}, y_{N + M})\}  $ with a total of $M$ samples.

\begin{figure}[htbp]
    \centering
    \includegraphics[scale = 0.45]{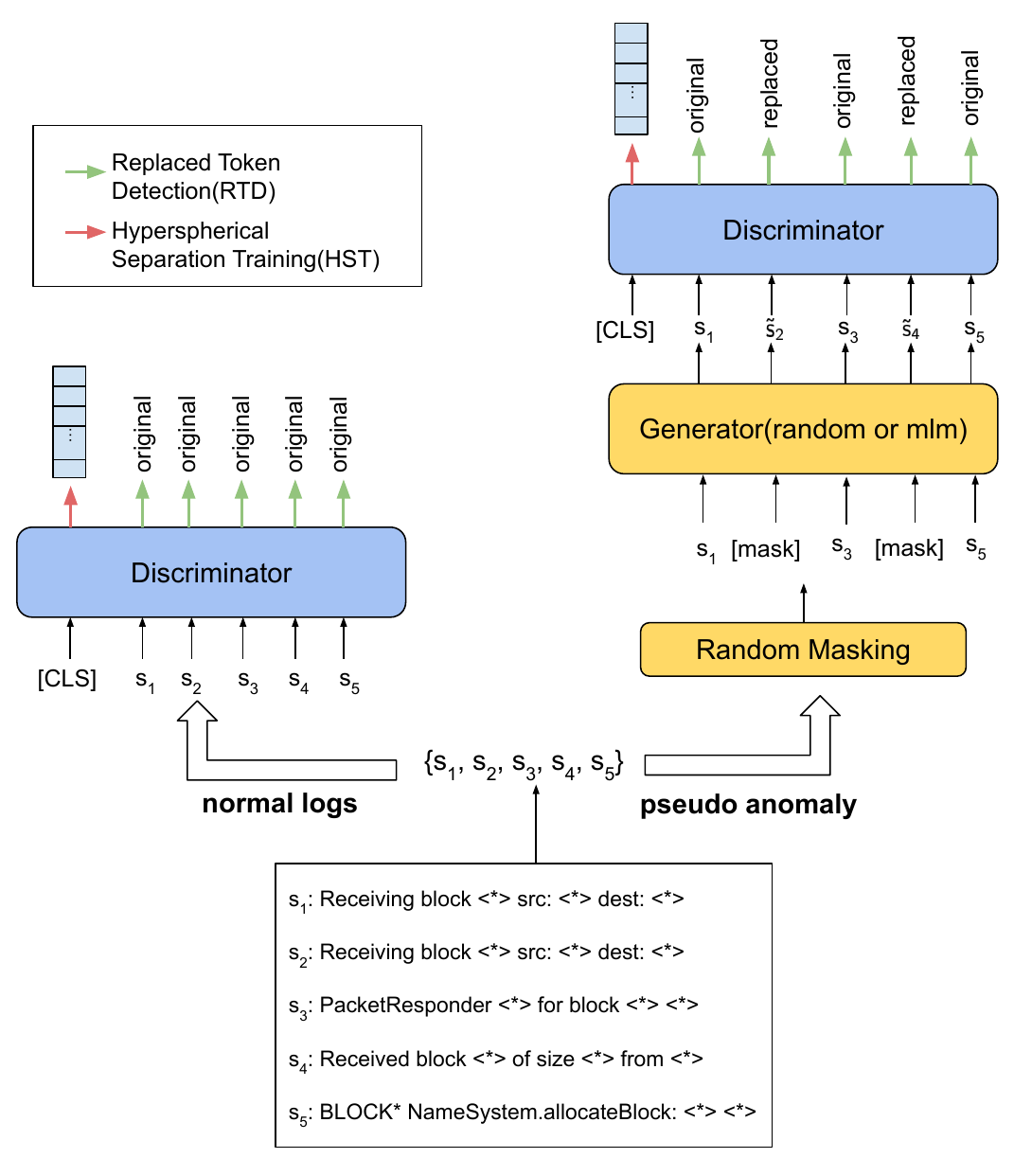}
    \caption{The training training procedure of FastLogAD. For a given sequence of normal logs, we randomly mask the log tokens in a certain ratio, and then generate the corresponding log sequence through a generator. For the discriminator, we propose RTD and HST to learn to distinguish normal logs from pseudo-anomaly logs.}
    \label{fig:training}
\end{figure}

\begin{figure}[htbp]
    \centering
    \includegraphics[scale = 0.45]{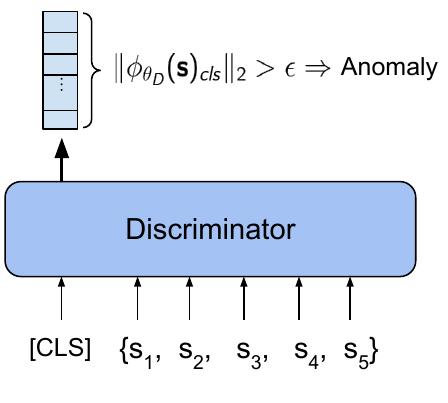}
    \caption{The inference procedure of FastLogAD. In inference, we directly use the anomaly discriminator for efficient diagnosis of logs.}
    \label{fig:testing}
\end{figure}

\textbf{Overview of FastLogAD model.} Fig.\ref{fig:training} and \ref{fig:testing} provide an overview of FastLogAD's training and inference stage, respectively. FastLogAD utilizes a generator-discriminator architecture that features Masked-Guided Anomaly Generation and Discriminative Abnormality Separation tasks. In this architecture, only the discriminator corresponding to Discriminative Abnormality Separation task is employed during inference. The generator's role is to generate pseudo-abnormal samples for each normal sequence during training. Two variants of the generator are introduced: the training-free random generator and the masked language model requiring training. The discriminator undergoes initial adversarial training to distinguish between normal and abnormal tokens replaced by the generator within a sequence, focusing on token-level recognition. Ultimately, it learns the output embedding to make normal and anomalous log sequences separable from a sequence-level recognition, following a similar approach presented in Logsy\cite{nedelkoski2020self}.

\subsection{Mask-Guided Anomaly Generation}
The Mask-Guided Anomaly Generation (MGAG) task in FastLogAD's framework serves to craft pseudo-abnormal log sequences for the discriminator's training. Given a normal log event sequence $\bm{s}=\{s_1, s_2, \dots, s_d\}$ from the training set, the tokens in the sequence undergo a random masking process with a masking ratio $r$, indicating what percentage of tokens gets replaced with $[mask]$ token. The resulting masking pattern $\bm{m}$ holding binary values indicates which tokens are replaced with the $[mask]$ token, and $\hat{\bm{s}} = \{\hat{s}_1, \hat{s}_2, \dots, \hat{s}_d\}$ denotes the masked sequence.
\begin{equation}
    \hat{s}_i = 
    \begin{cases}
        s_i & m_i = 0 \\
        [mask] & m_i = 1
    \end{cases}.
\end{equation}
By feeding the masked sequence to the generator, a pseudo-abnormal log sequence is constructed by replacing the masked tokens with random or unlikely tokens presented in a normal sequence. With this procedure, each training sample comes with corresponding abnormal versions of itself. Two variants of the generator are introduced in the following and using one of them can fulfill the MAGA task:
\begin{enumerate}
    \item The random generator that replaces the masked token with a random token chosen from the vocabulary. While this simple approach does not undergo any training process thereby reducing the computational overhead, it may fall short in accurately introducing the anomalous property of the whole sequence. This limitation arises because the random generator does not learn the dependencies among tokens within a sequence.
    \item In contrast to the random generator, the Masked Language Model (MLM) generator tends to replace masked tokens with those less likely to appear in the given sequence, crafting more realistic abnormal sequences. This generator exhibits improved performance, especially on datasets with a smaller vocabulary. However, this variant requires a training process and its effectiveness may diminish as the vocabulary size increases as shown in the following experimental results.
\end{enumerate}

\subsubsection{Random Generator}
As we create a vocabulary of tokens that represent the log events from the training set, the idea of the random generator is to replace the masked token with a randomly sampled token from the vocabulary. Let $\mathbf{V}$ represent the vocabulary, then the sequence generated by the random generator is denoted as $\tilde{\bm{s}}^{rand} = \{\tilde{s}^{rand}_{1}, \tilde{s}^{rand}_{2}, \dots, \tilde{s}^{rand}_{d}\}$, where
\begin{equation}
    \tilde{s}^{rand}_{i} =
    \begin{cases}
        s_i & m_i = 0 \\
        S \sim \mathbf{U}(\mathbf{V}\setminus\{s_i\}) & m_i = 1
    \end{cases}.
\end{equation}
Each $\tilde{s}^{rand}_{i}$ of masked input is sampled from a uniform distribution over the vocabulary $\mathbf{V}$ excluding $s_i$, the ground truth token from the original sequence.

\subsubsection{MLM Generator}
The Masked Language Model (MLM) generator uses a BERT with a classification head, outputting a probability distribution $P_G$ over the vocabulary $\mathbf{V}$. Unlike the random generator outputting a uniform distribution, this generator is trained using MLM loss:
\begin{equation}
    \mathcal{L}_{MLM} = -\mathbb{E}\left[\sum_{\substack{i = 1 \\ m_i = 1}}^d \log P_G(s_i \mid \hat{\bm{s}} ; \theta_G)\right].
\end{equation}
Although trained with this loss ensures the generator assigns higher probabilities to tokens likely to appear in the original normal sequence, our strategy for its output is to sample the tokens from the \textit{complement} distribution $\bar{P}_G$ such that for each $s_i \in \mathbf{V}$,
\begin{equation}
    \bar{P}_G(s_i \mid \hat{\bm{s}}; \theta_G) = \frac{1 - P_G(s_i \mid \hat{\bm{s}};\theta_G)}{\sum_{j = 1}^{|\mathbf{V}|}(1 - P_G(s_j \mid \hat{\bm{s}}; \theta_G))},
\end{equation}
where for each token the probability is normalized over the sum. Thus, the generated sequence by the MLM generator can be denoted as $\tilde{\bm{s}}^{mlm} = \{\tilde{s}^{mlm}_{1}, \tilde{s}^{mlm}_{2}, \dots, \tilde{s}^{mlm}_{d}\}$, where
\begin{equation}
    \tilde{s}_{mlm, i} = 
    \begin{cases}
        s_i & m_i = 0 \\
        S \sim \bar{P}_G(s \mid \hat{\bm{s}};\theta_G) & m_i = 1
    \end{cases}.
\end{equation}

\subsection{Discriminative Abnormality Separation}
The log sequence is prepended with a $[$CLS$]$ token before being fed to the discriminator. Such that, the input sequence is then $\tilde{\bm{s}} = \{\tilde{s}_{cls}, \tilde{s}_1, \tilde{s}_2, \dots \tilde{s}_d\}$ with $\{\tilde{s}_1, \tilde{s}_2, \dots, \tilde{s}_d\}$ being the given normal sequence with label $y = 0$ or the pseudo-abnormal version from the generator with label $y = 1$. In this Discriminative Abnormality Separation (DAS) task, we adopt a BERT \cite{devlin2018bert} model as the discriminator for achieving anomaly detection through two separate training tasks: Replaced Token Detection (RTD) and Hyperspherical Separation Training (HST).

\subsubsection{Replaced Token Detection (RTD)}
Similar to the MLM generator, the discriminator follows a BERT model structure connected to a binary classification for each token in the input sequence except for the added $[$CLS$]$ token. This classification head corresponds to the first training task, RTD being previously adopted effectively in ELECTRA's naive training \cite{clark2020electra} and its application DATE \cite{manolache2021date} of anomaly detection in NLP. In our approach, we apply this training as the warmup step for our discriminator to classify the replaced and non-replaced tokens for understanding the potential token-level anomalies before bringing them up to the overall anomaly detection on the whole sequence. Therefore, the training objective of RTD is to minimize the Binary Cross Entropy (BCE) loss across the training samples:
\begin{equation}
    \mathcal{L}_{RTD} = -\mathbb{E}\left[\sum_{i = 1}^d \log P_D(m_i \mid \tilde{\bm{s}}; \theta_D)\right]
\end{equation}

\subsubsection{Hyperspherical Separation Training (HST)}
In one-class anomaly detection modeling, we want the normal features tightly clustered within the latent space. To this end, we adopt a hyperspherical function in Logsy \cite{nedelkoski2020self} to separate normal and pseudo-abnormal sequences by controlling the norm of the feature embedding corresponding to the prepended $[$CLS$]$ token to the discriminator. Let $\phi_{\theta_D}(\tilde{\bm{s}})_{cls}$ denote this feature embedding, the loss of training this objective, namely Hyperspherical Separation Training (HST), is
\begin{align}
    \mathcal{L}_{HST} = &\mathbb{E}\Biggl[(1 - y)\|\phi_{\theta_D}(\tilde{\bm{s}})_{cls}\|_2 \nonumber \\
    &-\lambda y\log(1 - \exp (-\|\phi_{\theta_D}(\tilde{\bm{s}})_{cls}\|_2))\Biggl], \lambda > 0.
\end{align}
The first term enables normal samples, with the feature embedding norm minimized to be close to the origin. In contrast, the anomalous samples are enforced away from the origin as presented in the second term, penalizing the small embedding norm. This also prevents convergence to trivial solutions \cite{ruff2018deep}, where the parameters of the networks are zeros to minimize the first term only. $\lambda$ as a positive weight controls the emphasis between the two terms.

\subsection{Overall Training}
Our goal is to eventually detect anomalies based on the feature embedding norm of the $[$CLS$]$ token, as presented in the following section. Thus, we decide to employ a two-step training for our discriminator and generator (or training-free random generator):
\begin{itemize}
    \item Training the MLM generator together with the discriminator's RTD task for the first N epochs to start up: $\mathcal{L}_{MLM} + \mathcal{L}_{RTD}$ or $\mathcal{L}_{RTD}$ for the random generator. 
    \item M epochs for HST to promote accurate anomaly scores in the detection stage: $\mathcal{L}_{HST}$
\end{itemize}
This two-step training approach offers several advantages compared to joint training and separate training of all tasks:
\begin{enumerate}
    \item The joint training of RTD and MLM tasks for N epochs reduces training complexity, recognizing both tasks as auxiliary and not directly used in the subsequent detection stage. This streamlined approach avoids unnecessary computational overhead associated with separate training.
    \item RTD task is trained before HST to give the discriminator attention to the crafted abnormal tokens first for the subsequent learning of the sequence-level recognition. If joint training of RTD and HST is employed, the discriminator might overfit the crafted tokens instead of considering anomalies as entire sequences. Log anomalies are typically determined by the sequences they are part of, rather than individual log events. For instance, the HDFS dataset provides anomaly labels for entire grouped sequences rather than assigning a label to each log. Thus, RTD is exclusively trained before HST to guide the warm-up of training and later we show RTD training does help with our anomaly detection task.
\end{enumerate}

\subsection{Detection}
As the role of the generator is to craft the pseudo-anomalies for the training step, the anomaly detection for the incoming log sequences then only requires the discriminator to be performed. Each log sequence $\bm{s}$ combined with a leading $[$CLS$]$ token is directly fed to the discriminator. The discriminator evaluates each log sequence $\bm{s}$ based on the feature embedding norm $\|\phi_{\theta_D}(\bm{s})_{cls}\|_2$ as the anomaly score. Sequences are deemed abnormal if the score exceeds a threshold $\epsilon$, determined using a validation set comprising normal sequences.

\begin{table*}[htbp]
\sisetup{detect-all}
\NewDocumentCommand{\B}{}{\fontseries{b}\selectfont}
\hspace*{-1.1cm} 
\begin{tabular}{
  |l|
  l|
  l|
  l|
  l|
  l|
  l|
  l|
  l|
  l|
  l|
}
\hline
Method &
\multicolumn{3}{c|}{HDFS} &
\multicolumn{3}{c|}{BGL} &
\multicolumn{3}{c|}{Thunderbird} \\
\cline{2-10}
& {Precision} & {Recall} & {F1} & {Precision} & {Recall} & {F1} &{Precision} & {Recall} & {F1} \\
\hline
DeepLog \cite{du2017deeplog} & 79.37$\pm$4.24 & 92.06$\pm$0.28 & 85.23$\pm$3.67 & \underline{86.89$\pm$1.89} & 90.91$\pm$4.13 & 88.85$\pm$2.51 & 90.25$\pm$1.71 & 98.57$\pm$1.05 & 94.23$\pm$1.29\\
\hline
LogAnomaly \cite{meng2019loganomaly} & \B 93.46$\pm$2.67 & 59.20$\pm$4.55 & 72.48$\pm$5.08 & 78.03$\pm$1.51 & 86.70 $\pm$3.25 & 82.14$\pm$2.18 & 79.48$\pm$0.00 & 99.55$\pm$0.00 & 88.39$\pm$0.00\\
\hline
LogBERT \cite{guo2021logbert} &  \underline{85.20$\pm$0.58} & 75.39$\pm$1.29 & 80.00$\pm$0.68 & 74.79$\pm$0.96 & 91.68$\pm$1.45 & 82.38$\pm$0.78 & \B 96.27$\pm$0.40 & 95.03$\pm$2.23 & 95.65$\pm$1.14\\
\hline
FastLogAD-Random &83.56$\pm$0.75 & \underline{99.51$\pm$0.12} & \underline{90.84$\pm$0.40} & 82.72$\pm$3.14 & \underline{98.02$\pm$0.95} & \underline{89.72$\pm$1.49} & 94.88$\pm$0.12 &  \B{99.95$\pm$0.03} & \underline{97.34$\pm$0.02}\\			
\hline
FastLogAD-MLM & 84.80$\pm$0.60 & \B 99.99$\pm$0.01 & \B 91.77$\pm$0.36 & \B 88.28$\pm$3.58 & \B 98.66$\pm$0.86 & \B 93.14$\pm$1.66 & \underline{95.45$\pm$0.14} & \underline{99.91$\pm$0.08} & \B 97.63$\pm$0.03\\
\hline
\end{tabular}
\caption{Experimental results on HDFS, BGL and Thunderbird datasets, the best metric values are \textbf{bolded} and the runner up is highlighted with \underline{undeline}. Note that numerical results of the compared baselines are obtained with the best hyperparameters selected manually.}
\label{table:results}
\end{table*}

\begin{table*}[htbp]
\centering
\sisetup{detect-all}
\NewDocumentCommand{\B}{}{\fontseries{b}\selectfont}
\begin{tabular}{
  |l|
  S[table-format=3.2]|
  S[table-format=1]|
  S[table-format=3.2]|
  S[table-format=1]|
  S[table-format=3.2]|
  S[table-format=1]|
}
\hline
Method & 
\multicolumn{2}{c|}{HDFS} &
\multicolumn{2}{c|}{BGL} &
\multicolumn{2}{c|}{Thunderbird} \\
\cline{2-7}
& {Total (s)} & {Avg. (ms)} & {Total (s)} & {Avg. (ms)} & {Total (s)} & {Avg. (ms)} \\
\hline
DeepLog \cite{du2017deeplog} & 46.85 & 0.08 & 177.91 & 6.89 & 1647.97 & 14.14\\
\hline
LogAnomaly \cite{meng2019loganomaly} & 83.42 & 0.15 & 320.28 & 12.42 & 2953.37 & 25.34 \\
\hline
LogBERT \cite{guo2021logbert} & 791.12 & 1.39 & 132.05 & 5.11 & 936.40 & 8.04\\
\hline
FastLogAD-MLM & 77.32 & 0.14 & 18.57 & 0.72 & 78.06 & 0.67\\
\hline
\end{tabular}
\caption{Total inference time and the average inference time per log sequence.}
\label{table:inference time}
\end{table*}

\section{Experiment}
\subsection{Experimental Settings}
\subsubsection{Datasets}
\begin{itemize}[parsep=0pt, topsep=0pt]
    \item \textbf{HDFS} \cite{xu2009detecting} logs were collected from a Hadoop cluster running on over 200 EC2 nodes, yielding 11,175,629 lines of logs with $block\_id$ as the identifier for grouping. Among the total of 575,061 blocks of logs in the HDFS dataset, 16,838 blocks were labelled as anomalies.
    \item \textbf{BGL} \cite{oliner2007supercomputers} was collected from the Blue Gene/L supercomputer deploying at Lawrence Livermore National Laboratory (LLNL). It contains 4,747,963 logs and 348,460 logs were manually labelled as anomalies. Different from HDFS having an identifier, we group BGL logs using a 5-minute sliding window with a 1-minute step size, following the grouping strategy applied in LogBERT \cite{guo2021logbert}.
    \item \textbf{Thunderbird} \cite{oliner2007supercomputers} consists of system logs (syslog) in a total of 211,212,192 collected from Sandia National Labs (SNL). To shrink the dataset for efficiency in training and inference, we take the first 20M logs containing 758,562 anomalous logs from the original dataset and again following LogBERT \cite{guo2021logbert}, we group the selected logs using a 1-minute sliding window with a 30-second step size.
\end{itemize}
For all datasets, a chronological order is maintained during the train-test split, ensuring that normal sequences are split without shuffling. Specifically, we follow the experiment setting in LogBERT \cite{guo2021logbert} and take around the first 5000 log sequences following the given timestamp as training data to reflect practical scenarios. Subsequently, 10\% of the split training data is set aside for validation, and all abnormal sequences are included in the test set for evaluation.

\subsubsection{Baselines} 
Due to the limited available public implementations \cite{le2022log} of unsupervised log anomaly detection with deep learning, we compare our FastLogAD to log anomaly detection baseline models including DeepLog \cite{du2017deeplog}, LogAnomaly \cite{meng2019loganomaly} and LogBERT \cite{guo2021logbert}. In addition, we also present additional comparison studies in Appendix. 

\subsubsection{Evaluation Metrics}
Following the existing works, we adopt Precision, Recall and F1-Score as evaluation metrics:
\begin{align*}
    \text{Precision} &= \frac{\text{TP}}{\text{TP} + \text{FP}} ; \text{Recall} = \frac{\text{TP}}{\text{TP}+\text{FN}}; \\
    \text{F1-Score} &= 2\times\frac{\text{Precision}\times\text{Recall}}{\text{Precision} + \text{Recall}}  
    \intertext{TP: True Positive; FP: False Positive; FN: False Negative;} \nonumber
\end{align*} 

\subsubsection{Experimental Setup}
We conduct our experiments on a Linux Server with Intel i9-9940X @ 3.30GHz CPU and 64GB of RAM. One Nvidia RTX 2080Ti GPU is utilized for training under the Python 3.8.10 environment with Pytorch 2.1.0+cu118 installed.

\subsubsection{Implementation Details}
For FastLogAD, we take $r = 0.5$ mask ratio, $\lambda = 1$ for HST loss. The detection threshold $\epsilon$ is determined by taking the 99\% quantile of embedding norms on the validation dataset to exclude the outliers. We adopt the architecture of ELECTRA \cite{clark2020electra} to construct both the MLM generator and discriminator for both variants. The single embedding layer is configured with a dimension of 256, followed by 4 transformer layers. Each transformer layer is equipped with 4 heads, and the feedforward layer has a dimension of 256. More detailed parameters can be found in our public implementations. For the baselines, DeepLog \cite{du2017deeplog} and LogAnomaly \cite{meng2019loganomaly} do not have their official implementations available to the public, therefore we experimented with others' replicated versions\footnote{\url{https://github.com/donglee-afar/logdeep}, \url{https://github.com/LogIntelligence/LogADEmpirical}} instead. All the involved hyperparameters are selected according to the adopted implementations and the detection thresholds of these three baselines: $k$ presented in DeepLog \cite{du2017deeplog} and LogAnomaly \cite{meng2019loganomaly} and anomalous tokens ratio $r$ in LogBERT \cite{guo2021logbert} are selected according to their best performance on the test set since these thresholds need to be determined with abnormal sequences involved. While this provides a natural advantage for these baselines in comparing performances to FastLogAD, the impracticality of selecting thresholds should be considered. 

\subsection{Evaluation}
Table \ref{table:results} shows the experimental results of the baselines and FastLogAD with both two generator variants. The results are based on running with 3 random seeds and all metric values are expressed as percentages (\%). FastLogAD outperforms the compared baselines in terms of F1-Score. Specifically, FastLogAD with the MLM generator achieves the best F1-Score across three datasets, while the random generator variant achieves the second best, with a slight lag behind the MLM generator approach on Thunderbird. This slight lag is caused by the large vocabulary size ($\approx$1000) of the training data in which training MLM is nearly equivalent to training a random generator by eventually sampling from a large number of candidates. In an overall comparison with the best-performing baselines on the three datasets, FastLogAD exhibits a 6.54\% and 4.29\% advantage in F1-Scores over DeepLog for HDFS and BGL, respectively. Concerning the Thunderbird dataset, FastLogAD achieves a 1.98\% higher F1-Score than LogBERT.

\subsubsection{Inference Time Analysis}
As we discussed in Introduction, the speed of log anomaly detection plays a pivotal role in the realm of system maintenance and security. so in this experiment, we evaluate the inference time of FastLogAD and other competitors to demonstrate the effectiveness of our proposed method. As shown in Table \ref{table:inference time}, we present the total inference time and average inference time per log sequence for the compared models across three datasets. LogBERT \cite{guo2021logbert} achieves slower inference on the HDFS dataset, which is composed of short sequences. This is caused by the random masking step and multiple predictions performed on each sequence. DeepLog \cite{du2017deeplog} and LogAnomaly \cite{meng2019loganomaly} have the best efficiency with HDFS dataset, of which the log sequences are relatively shorter than the other two datasets. Their inference speed noticeably deteriorates with long sequences where each sequence needs to be further decomposed into multiple short sequences to maintain the performance. In contrast, FastLogAD showcases satisfactory overall inference speed across datasets. Its practicality for real-time inference is noteworthy, as it involves only the discriminator, requiring no additional operations on the input sequence. This characteristic makes FastLogAD suitable for both long and short-sequence datasets, presenting a distinct advantage over prior arts.
\begin{figure*}[htbp]
    \centering
    \includegraphics[width=0.27\textwidth]{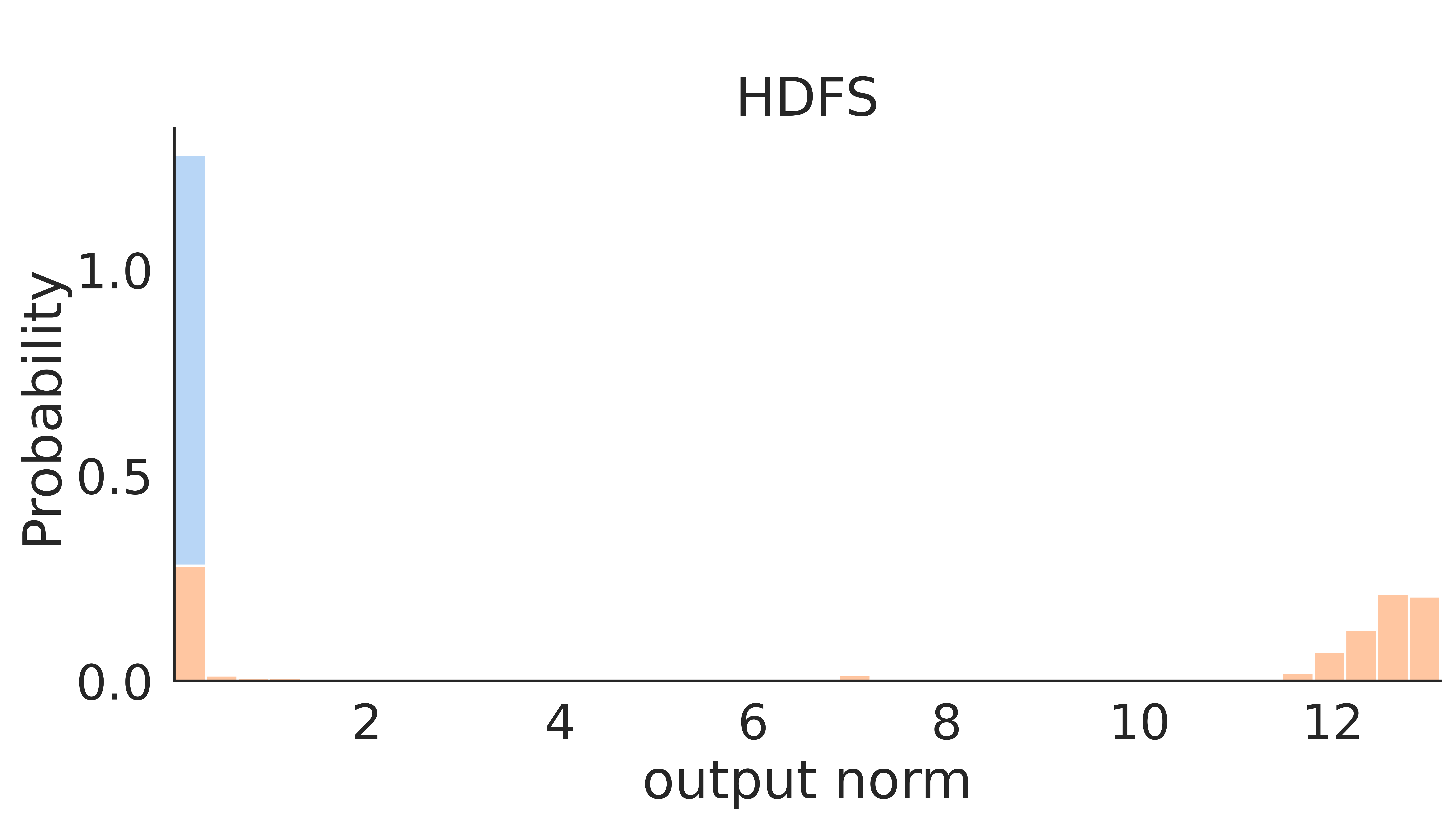}
    \includegraphics[width=0.27\textwidth]{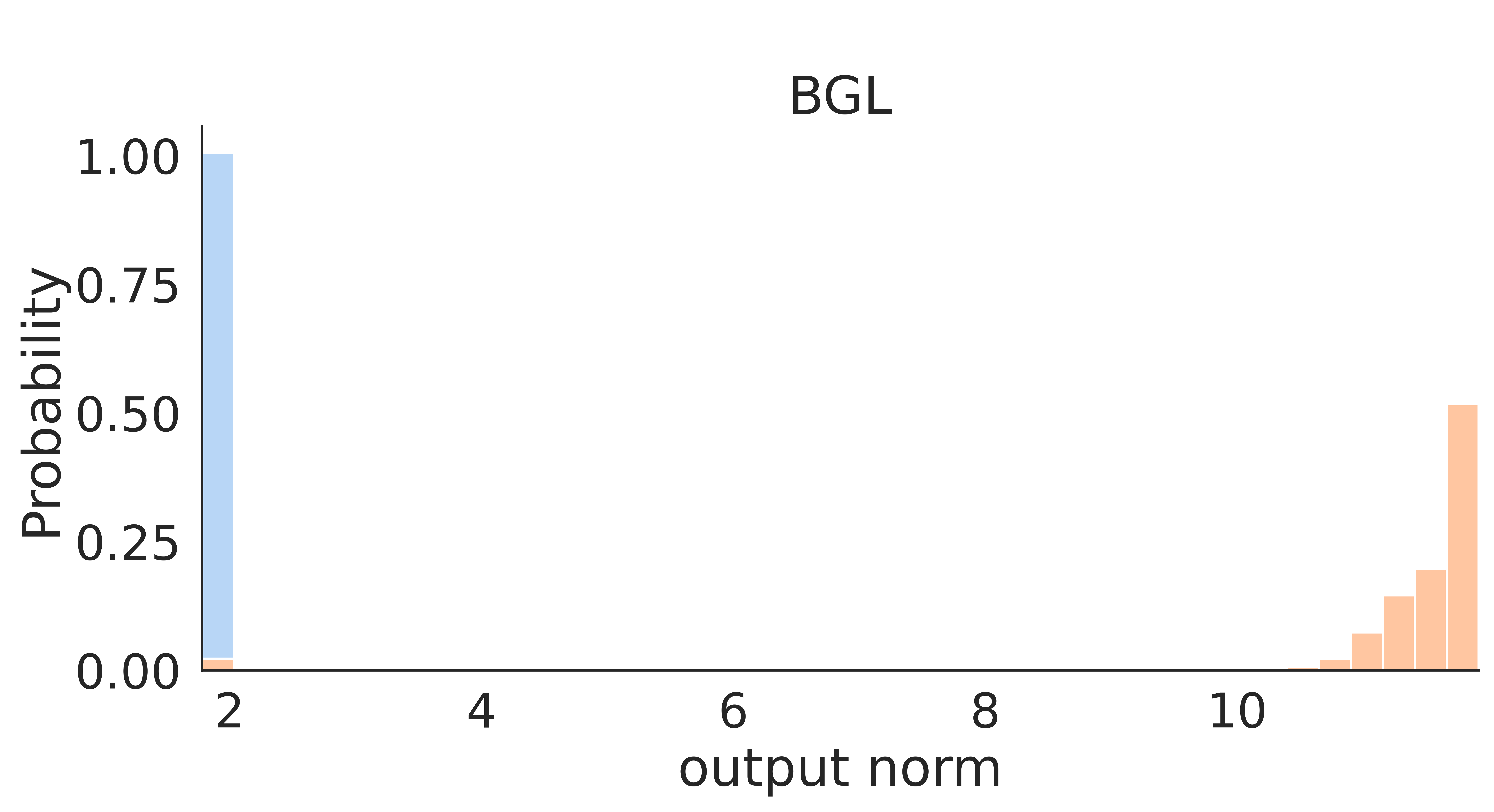}
    \includegraphics[width=0.37\textwidth]{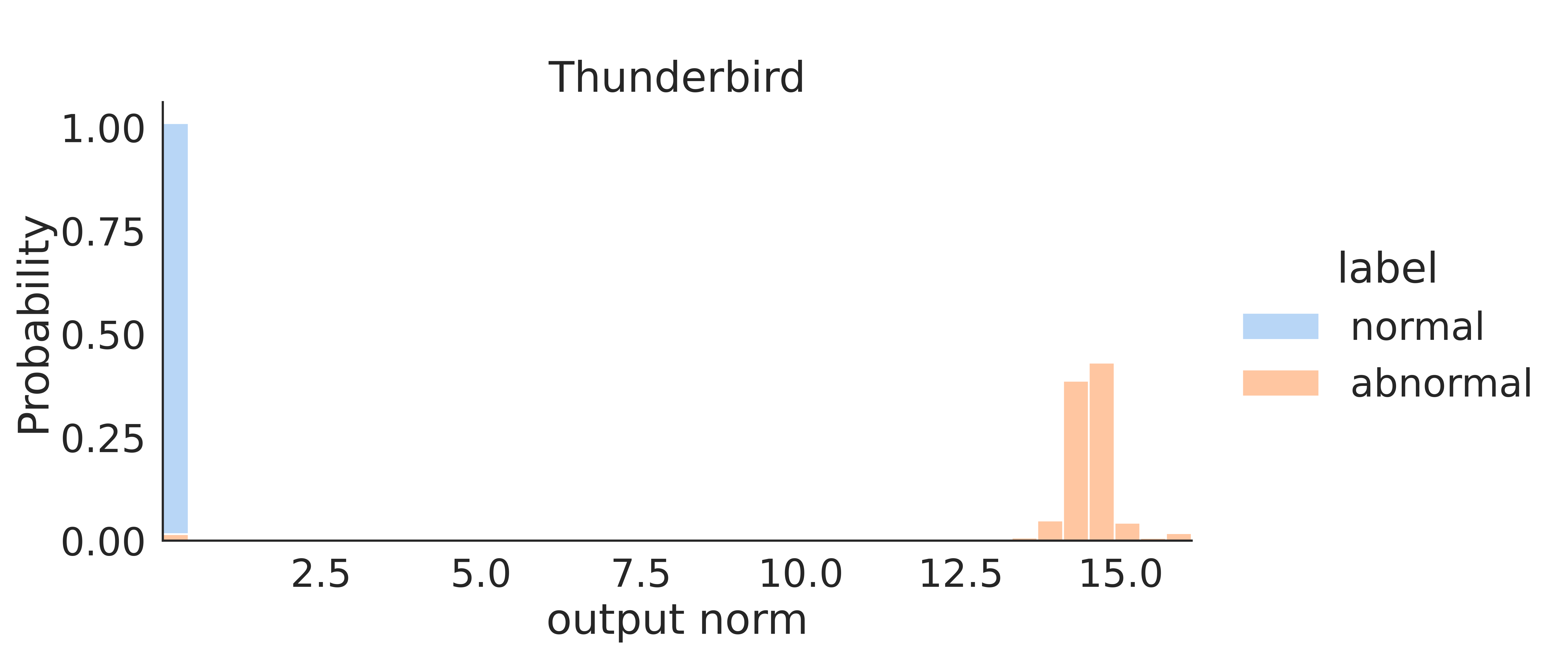}
    \caption{Visualization of anomaly probability distributions on HDFS, BGL and Thunderbird datasets.}
    \label{fig:distplot}
\end{figure*}
\begin{figure*}[htbp]
    \centering
    \includegraphics[scale = 0.27]{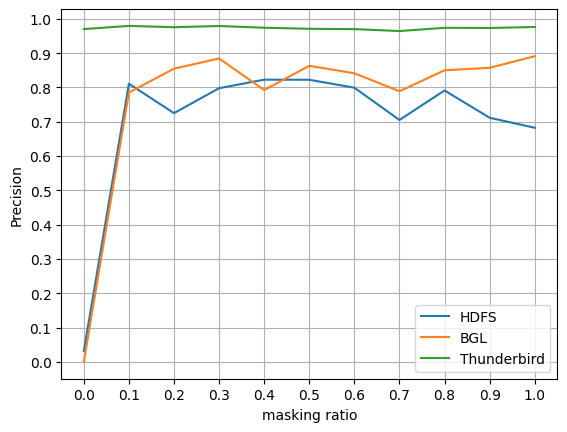}
    \includegraphics[scale = 0.27]{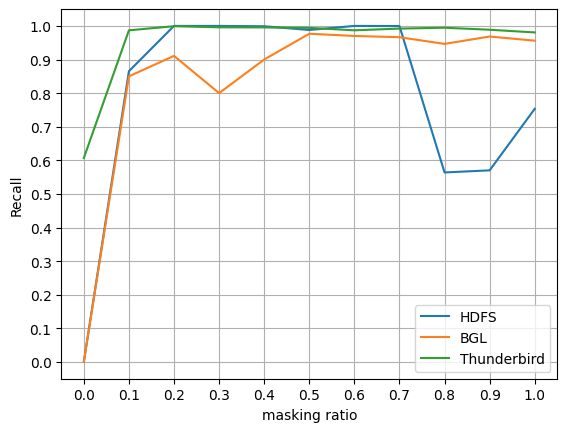}
    \includegraphics[scale = 0.27]{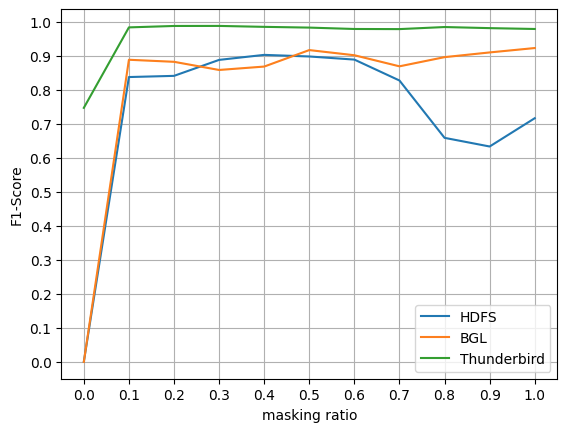}
    \caption{Performance comparison of FastLogAD-MLM across different masking ratios (0 to 1).}
    \label{fig:masking_ratio}
\end{figure*}
\subsubsection{Distribution of Anomaly Score}
Fig. \ref{fig:distplot} displays the distribution plot of anomaly scores for the three experimental datasets. The anomaly scores of normal and abnormal data are depicted in distinct colours, showing that almost exclusively normal data is concentrated near the origin. Moreover, a significant portion of abnormal data has an anomaly score exceeding 10. This substantial separation ensures a distinct separation between normal and abnormal data, contributing to FastLogAD-MLM's competitive performance. Notably, this achievement is noteworthy as the anomaly threshold is selected without prior exposure to abnormal data. 

\subsubsection{Additional Evaluation}
Our main experiment utilizes a small ratio of training samples for the sake of practicability and emphasis on FastLogAD's competitive performance. Though there is no standard way of splitting the log dataset, we additionally evaluated FastLogAD with a split ratio of 6:1:3 on the BGL dataset following the strategy introduced in GLAD \cite{li2023glad}. Due to the missing files in the author's provided code of GLAD \cite{li2023glad}, we are unable to directly compare with it in the main experiment. In this additional experiment, the results of the compared methods are adopted from the GLAD \cite{li2023glad} paper. We append FastLogAD's results to them, all together displayed in the \textbf{Appendix}.

\subsection{Ablation Studies}
\subsubsection{RTD training}
To assess the impact of RTD training, we report the performance of FastLogAD-MLM and FastLogAD-Random without RTD training in Table \ref{table:RTD}. A comparison with the original FastLogADs demonstrates that RTD training proves effective, resulting in slightly improved performance across all three datasets. This enhancement can be attributed to RTD's support of the subsequent HST task.
\begin{table}[htbp]
\centering
\sisetup{detect-all}
\NewDocumentCommand{\B}{}{\fontseries{b}\selectfont}
\begin{tabular}{
  |l|
  p{0.8cm}|
  p{1.6cm}|
  p{0.8cm}|
  p{1.6cm}|
}
\hline
Backbone &
\multicolumn{2}{c|}{Random} &
\multicolumn{2}{c|}{MLM} \\
\hline
Setting & {w/o RTD} & {w/ RTD} & {w/o RTD} & {w/ RTD} \\
\hline
HDFS & 90.01 & 90.55 (+0.54) & 90.70 & 90.76 (+0.06) \\

BGL & 88.34 & 89.58 (+1.24) & 91.50 & 91.64 (+0.14) \\

Thunderbird & 98.24 & 98.72 (+0.48) & 98.42 & 98.61 (+0.19) \\
\hline
\end{tabular}
\caption{Ablation studies in terms of F1-score. We denote w/ and w/o as with RTD and without RTD, respectively.}
\label{table:RTD}
\end{table}

\subsubsection{Masking ratio}
Additionally, we explore the influence of the masking ratio on FastLogAD's performance. Experimenting with a masking ratio ranging from 0 to 1, incremented by 0.1, we train FastLogAD-MLM on each dataset. Fig. \ref{fig:masking_ratio} with plotted three evaluation metrics reveals a significant drop in performance at $r = 0$, as intact normal sequences are treated as pseudo-anomalies, leading the model to inflate the embedding norm of normal data. However, for $r > 0$, FastLogAD demonstrates effectiveness in creating pseudo-anomalies. A notable observation is a performance drop at higher masking ratios, starting from 0.7 on the HDFS dataset. This can be attributed to the fact that the HDFS dataset comprises relatively short log sequences, making the distinction between normal and abnormal sequences less obvious when considering all the tokens. This implies with a large masking ratio, the model may capture only the patterns of easy pseudo-anomalies, aligning with real anomalies in datasets with long log sequences like BGL and Thunderbird. However, these easy pseudo-anomalies may not align with the characteristics of the HDFS dataset, causing the model to misclassify real hard anomalies as normal. This corresponds to the increasing number of false negatives and a large drop in recall values as shown in the plot.

\section{Conclusion and Future Work}
In this study, we introduced FastLogAD, a novel approach for unsupervised log anomaly detection that leverages a generator-discriminator architecture inspired by ELECTRA. By training on labelled normal data and generating pseudo-abnormal samples, FastLogAD enriches the training set thereby strengthening the discriminator's ability to detect anomalies effectively. While promising, future work should explore scalability and adaptability to various log formats, ensuring the model's relevance in evolving log data scenarios and real-time deployment environments.

While FastLogAD has demonstrated superior performance compared to all included baselines, opportunities for improvement are still evident. In this section, we outline potential directions for future research:
\begin{itemize}
    \item Despite its overall high F1-Score, FastLogAD exhibits a comparatively lower precision score. This discrepancy is attributed to false positive predictions, particularly when sequences from the test set are out of the vocabulary trained based on the training data. Such sequences are encoded into $[unk]$ tokens, increasing the likelihood of being recognized as anomalies by FastLogAD. To address this, future research should explore the incorporation of additional features, such as semantic features, to enhance robustness against Out-Of-Vocabulary (OOV) issues. Also, crafting corresponding features for pseudo-anomalies is necessary for improved precision.
    \item Currently designed for unsupervised anomaly detection, FastLogAD could be extended to support pure unsupervised anomaly detection. This adaptation would eliminate the need for fully normal training data, consequently reducing the cost associated with manual labelling and broadening FastLogAD's applicability.
\end{itemize}



\bibliographystyle{ACM-Reference-Format}
\bibliography{bibliography}

\appendix
\twocolumn
\section{Additional Experiment}
This section shows the additional experiment on the BGL dataset with a split ratio of 6:1:3. Both the evaluation scores and computational overhead of FastLogAD are included respectively in two tables.  The results of other baselines are adopted from the paper of GLAD \cite{li2023glad} and we aim to evaluate them independently once the issue in the public implementation of GLAD \cite{li2023glad} is fixed.
\subsection{Evaluation Performance}
\begin{table}[htbp]
\hspace*{-0.4cm}
\centering
\sisetup{detect-all}
\NewDocumentCommand{\B}{}{\fontseries{b}\selectfont}
\begin{tabular}{
  |l|
  S[table-format=2.2]|
  S[table-format=2.2]|
  S[table-format=2.2]|
  S[table-format=2.2]|
  S[table-format=2.2]|
}
\hline
{Method} & {Precision} & {Recall} & {F1} & {AUC} & {AUPR} \\
\hline
DeepLog \cite{du2017deeplog} & 89.02 & 80.54 & 84.57 & 89.26 & 70.17\\
\hline
LogAnomaly \cite{meng2019loganomaly} & 91.40 & 79.32 & 84.93 & 92.98 & 75.21\\
\hline
LogBERT \cite{guo2021logbert} &  91.47 & 92.69 & 92.07 & 96.33 & 82.70\\
\hline
LogGD \cite{xie2022loggd} & 90.89 & 93.31 & 92.08 & 96.91 & 81.74 \\
\hline
LogFlash \cite{jia2021logflash} & 82.46 & 86.73 & 84.54 & 86.78 & 74.52\\
\hline
DeepTraLog \cite{zhang2022deeptralog} & 79.48 & \B 97.68 & 87.64 & 84.77 & 70.92\\
\hline
GLAD \cite{li2023glad} & 90.82 & 94.57 & 92.66 & \B 98.18 & 84.69\\
\hline
FastLogAD-Random & 94.11 & 95.00 & 94.50 & 97.22 & \B 95.47 \\	
\hline
FastLogAD-MLM & \B 95.92 & 95.20 & \B 95.55 & 97.42 & 95.39 \\
\hline
\end{tabular}
\caption{Additional evaluation scores on BGL dataset, the best score under each metric is bolded.}
\label{table:additional score}
\end{table}
Table \ref{table:additional score} presents the evaluation scores. Compared to the baselines including the additional GNN approaches, our FastLogAD-MLM can achieve the best F1-Score under this new setting, with 1\% advantage over GLAD \cite{li2023glad}. The threshold selected for anomaly detection is still the 99\% quantile of the embedding norm of the validation set. In terms of the AUC and AUPR metrics, both variants of FastLogAD hold a superior overall score, highlighting our model's discriminability under various thresholds. They also exhibit a large lead in AUPR, with over 10\% advantage over the best baseline, showing exceptional performance on an imbalanced log dataset.

\end{document}